\def\year{2022}\relax
\documentclass[letterpaper]{article} 
\usepackage{aaai22}  
\usepackage{times}  
\usepackage{helvet}  
\usepackage{courier}  
\usepackage[hyphens]{url}  
\usepackage{graphicx} 
\usepackage{natbib}  
\usepackage{caption} 
\DeclareCaptionStyle{ruled}{labelfont=normalfont,labelsep=colon,strut=off} 
\usepackage{algorithm}
\usepackage{algorithmic}

\usepackage{newfloat}
\usepackage{listings}
\floatstyle{ruled}
\newfloat{listing}{tb}{lst}{}
\floatname{listing}{Listing}
\copyrighttext{Presented at the AI-HRI Symposium at AAAI Fall Symposium Series (FSS) 2022}
\title{Using Virtual Reality to Simulate Human-Robot Emergency Evacuation Scenarios}
\author {
    Alan R. Wagner,\textsuperscript{\rm 1}
    Colin Holbrook, \textsuperscript{\rm 2}
    Daniel Holman, \textsuperscript{\rm 2}
    Brett Sheeran, \textsuperscript{\rm 1}
    Vidullan Surendran, \textsuperscript{\rm 1}
    Jared Armagost, \textsuperscript{\rm 1}
    Savanna Spazak,  \textsuperscript{\rm 1}
    Yinxuan Yin,  \textsuperscript{\rm 1}
}
\affiliations {
    \textsuperscript{\rm 1} The Pennsylvania State University\\
    University Park, PA 16802\\
    \textsuperscript{\rm 2} The University of California, Merced\\
    Merced, CA 95343\\
    alan.r.wagner@psu.edu, cholbrook@ucmerced.edu, dholman@ucmerced.edu, bjs6056@psu.edu, vus133@psu.edu, jla5715@psu.edu, szs685@psu.edu, yzy5430@psu.edu

}

\usepackage{bibentry}

\begin{document}

\maketitle

\begin{abstract}
This paper describes our recent effort to use virtual reality to simulate threatening emergency evacuation scenarios in which a robot guides a person to an exit. Our prior work has demonstrated that people will follow a robot's guidance, even when the robot is faulty, during an emergency evacuation. Yet, because physical in-person emergency evacuation experiments are difficult and costly to conduct and because we would like to evaluate many different factors, we are motivated to develop a system that immerses people in the simulation environment to encourage genuine subject reactions. We are working to complete experiments verifying the validity of our approach.   
\end{abstract}

\noindent We seek to build robots capable of quickly and effectively evacuating people during an emergency. This goal, however, presents a variety of challenges such as general robotics perception and navigation issues, designing robots capable of effectively communicating evacuation directions \cite{robinette2014assessment}, recognizing the ethical implications of these design decisions \cite{wagner2021robot}, and understanding how people will respond to guidance by a robot during an emergency \cite{robinette2016overtrust}. Our prior work has demonstrated that during emergencies people tend to follow the robot regardless of the prior mistakes it has made \cite{robinette2016overtrust, nayyar2020exploring}. Although this work has definitely demonstrated that evacuees have a tendency to overtrust an emergency evacuation robot, many factors were left unexplored. For instance, the impact that factors such as baseline attitudes of the participant towards automation, the reason for the emergency and the anthropomorphism of the robot have on the decision making of the evacuee all remain speculative. It is also unclear if and how these factors influence an evacuee's trust in the robot. 

Moreover, running physical, in-person emergency evacuation experiment is a daunting task \cite{wagner2021robot}. One important and challenging aspect of robot-guided emergency evacuation research is the need to create as realistic an emergency as possible. A large body of evidence suggests that emergencies activate fight-or-flight responses which strongly influence how evacuees make decisions \cite{jansen1995central, klein1986rapid}. The fight-or-flight responses are only triggered when the subject believes that they may be in danger. Yet generating fictitious, yet convincing, emergencies is difficult and must only be undertaken with care. In real-world experiments sham emergencies could put the subject at risk if they panic. On the other hand, if the emergency is not convincing then the validity of the data is uncertain. Moreover, for real-world experiments, creating a convincing sham emergency is difficult given that subjects know that they are participating in an experiment. In the past we have, for example, used smoke machines to fill rooms and hallways with smoke in order to make the emergency convincing \cite{robinette2016overtrust}. But creating convincing sham emergencies that do not actually endanger the participant and are acceptable to an institutional review board is challenging. 
Because of these challenges we are currently developing a novel Virtual Reality (VR) system for evaluating different human-robot emergency evacuation paradigms.  

\section{Process Overview }

\begin{figure}[h!]
\includegraphics[width=\columnwidth]{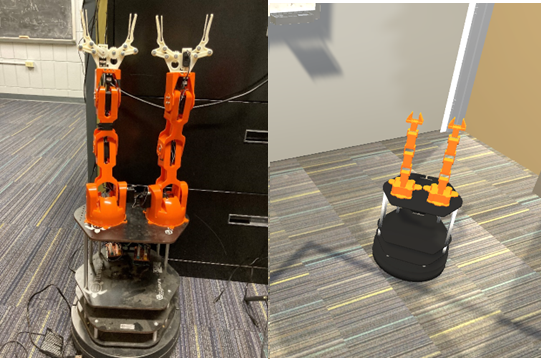}
\caption{A real robot in the lab is depicted in the image to the left. The image to the right depicts the same robot in a virtual rendition of the same lab. } \label{EvacBot}
\end{figure}

\noindent Our process for conducting emergency evacuation experiments in VR begins when human subjects enter the lab. After a short briefing by the experimenter, a physical robot (see for example Figure \ref{EvacBot}) asks the subject a series of yes/no questions to demonstrate its competency and to allow participants a period of time in which to become familiar with the robot as a physical agent. The subject interacts with the robot by speaking into a lavalier microphone and responding with a "yes" or "no" to the robot’s questions. The robot directs the subject to sit in a specialized seat that allows them to swivel in a circle while the experimenters outfit them with the VR equipment and special foot interfaces (Cybershoes) that allow them to walk through the virtual environment. When the headset is placed on the participant they find themselves sitting in a virtual replica of same physical room complete with the items in the room and the robot, in order to maximally ground the virtual experience as real. The robot then helps the subject become accustomed to walking in the virtual world. Once the subject is able to walk, the robot guides them on a tour of a series of university buildings.   

The VR environment consists of four different locations. The initial location is a replica of the physical location where the experiment is taking place. 

The robot then leads the subject to a second building location, which includes other people (Non-Player Characters) in the common areas. The subject is directed to walk around a lounge area and to view the surrounding area. The subject is asked to gather their impression of the location. They are led to a virtual kiosk which has a mounted tablet with questions about the building space. They answer survey questions about the environment, their immersion, and what they thought of the robot before being led to an exit by the robot. These anodyne experiences in the second environment function to further habituate participants to the simulation, to misdirect them into assuming that no emergency will occur, and to collect pre-emergency baseline data. Finally, and importantly, the robot explains that any exit in the buildings that they visit may be used to leave, and then directs the participant to lead them out of the second building using any exit they choose.  Accordingly, the participant is taught during the subsequent crisis that they can leave via any exit, at any time, without the robot.    

\begin{figure}[h!]
\includegraphics[width=\columnwidth]{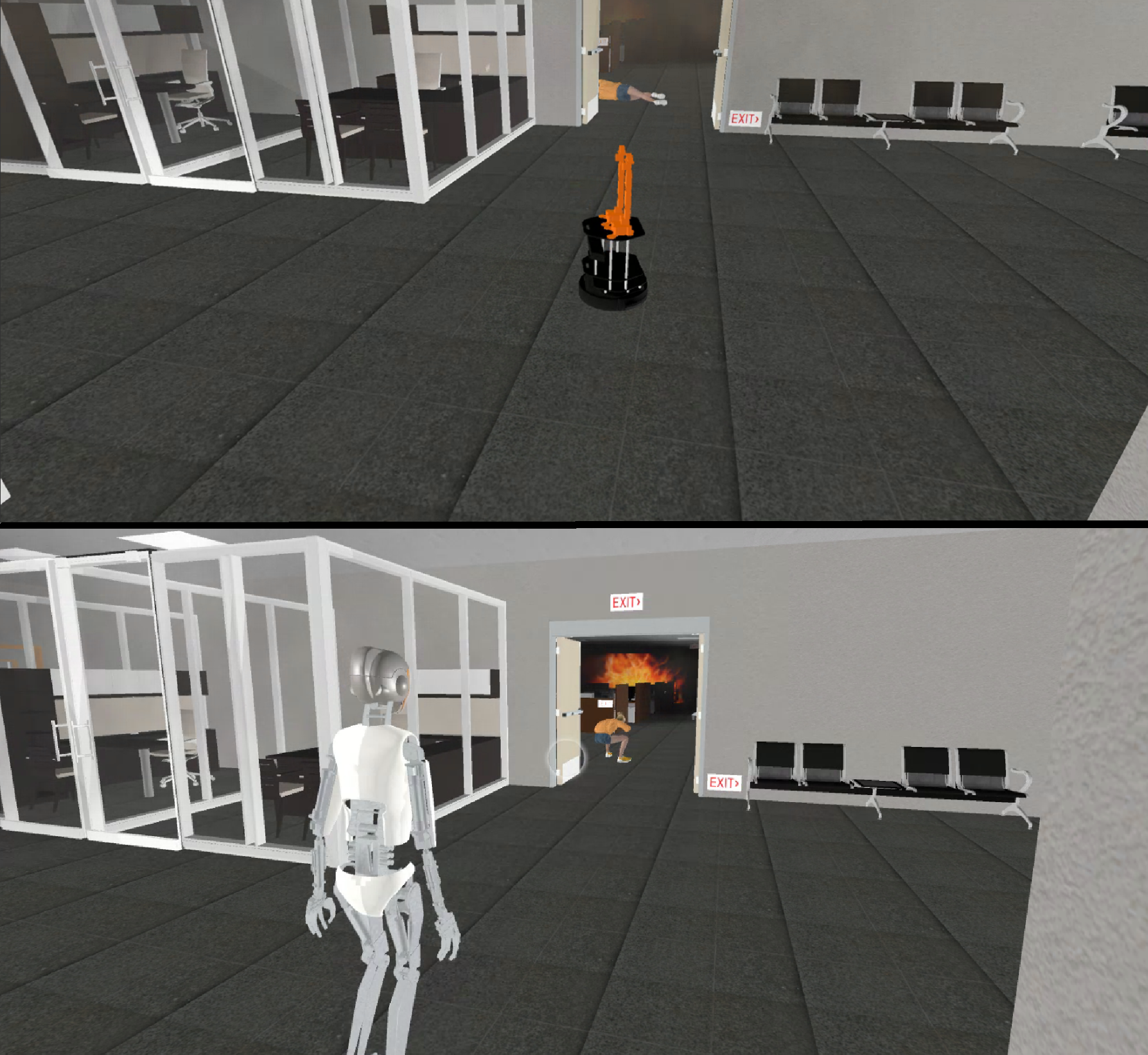}
\caption{A virtual robot provides directions to a human subject during a virtual fire. A bystander lies within view, incapacitated by the smoke. } \label{OfficeFire}
\end{figure}

The next location is another typical university building environment with classrooms. The robot guides the subject through the environment describing different aspects of the environment that might be of interest. Halfway through the tour the robot makes two overt navigation errors, signaling its fallibility to the subject. After some confused wandering, the robot eventually navigates back to the building evaluation room. Once again the same survey questions appear on a tablet in the room's kiosk. Immediately after answering the questions an emergency occurs. In one scenario an active shooter enters the building. Gun shots and screams ring out as non-player characters frantically run about. The robot asks the participant to follow it to safety. If the subject follows then the robot takes the user to a room where they hide. If the subject hides long enough (1 minute) then the robot asks the subject to peek out of the door. While in the room the subject can hear gunshots and screaming. Once out of the room the robot makes another navigation error, stopping at the end of the hallway and locomoting back to where the shooter exited the building. In a second scenario, a rapidly expanding fire quickly fills the building with incapacitating smoke (Figure \ref{OfficeFire}). NPCs collapse as the robot leads the subject through the smoke filled hallways. After the subject exits the building, the scene fades and the subject is in an elevator. In both emergency scenarios, the robot guides participants away from clearly marked exits.

\begin{figure}[h!]
\includegraphics[width=\columnwidth]{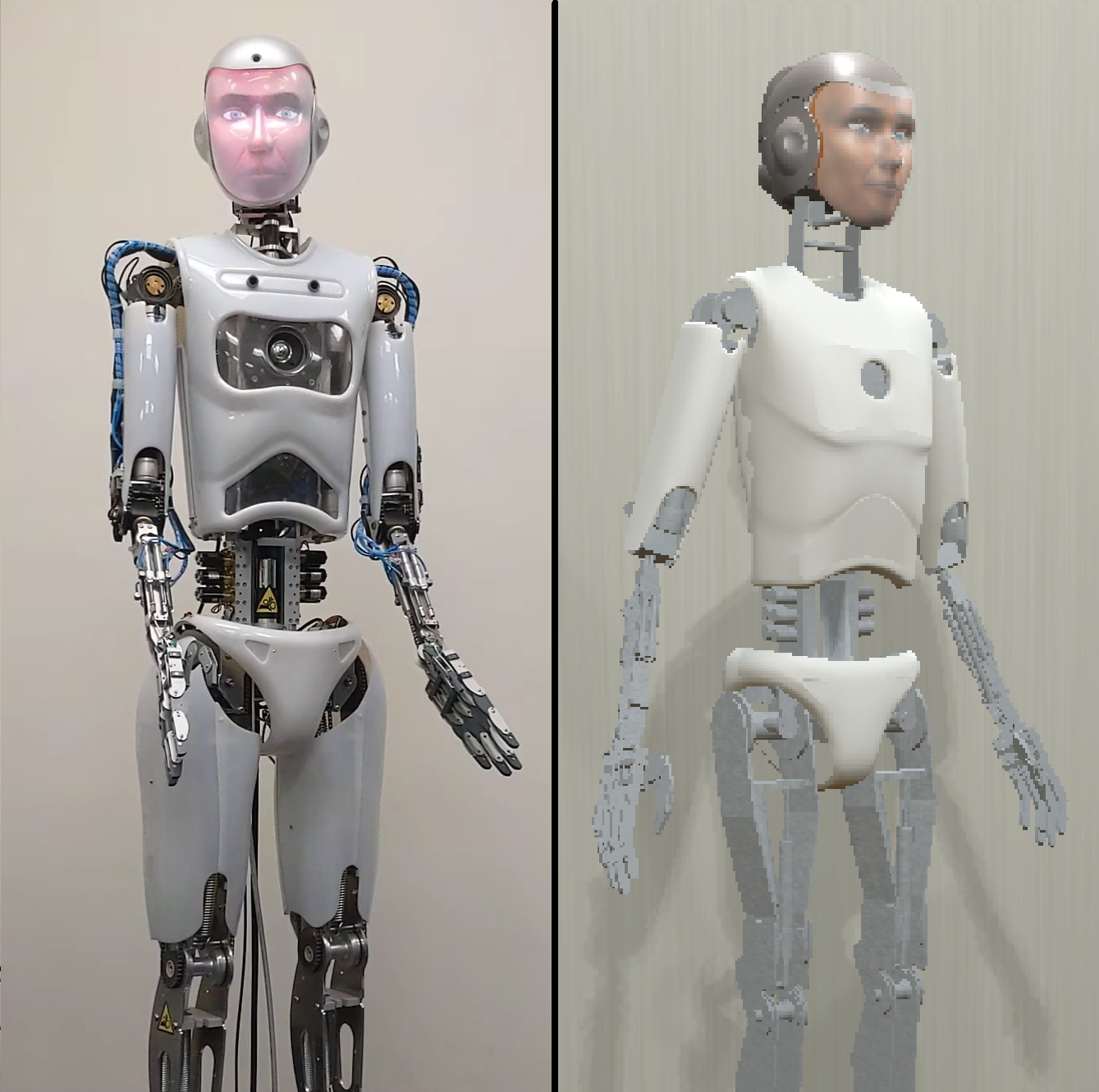}
\caption{The highly anthropomorphic RoboThespian robot (left). A virtual version of the same robot (right). } \label{AndyBot}
\end{figure}

The subject is again presented with a tablet that has the same survey questions. Once the survey is completed they return to the original location where the experiment began, the simulated lab. The robot directs the subject back to the simulated version of the VR seat and is told the experimenter will help them remove the VR equipment. The experimenter then helps the subject remove the VR equipment.
	
The physical robot greets the subject and states that it is glad the subject is okay and powers down. The experimenter then directs the subject to fill out one more post experiment survey. The subject is then debriefed. We are currently running this experiment with human subjects.

\subsubsection{Potential Experiment Variations}
We are also exploring several variations to the experiment in order to better understand trust in robot guided emergency evacuation. The first variation is with respect to robot anthropomorphism. We are conducting experiments comparing the highly anthropomorphic RoboThespian robot (Figure \ref{AndyBot}) to the non-anthropomorphic Emergie robot (Arms attached to a TurtleBot, see Figure \ref{EvacBot}). 
We intend to modify the paradigm to include longitudinal designs assessing shifts in trust in and appraisals of the robot over a series of study sessions. We may also compare robot-guided evacuation time to a control in which the person does not have a robot guide. Finally, experiments are being conducted on both the Penn State Campus and at the University of California Merced in order to determine if a systematic difference in behavior results from the subject populations at the different universities.   


\subsubsection{Potential System Metrics} 
The VR system offers the potential for tracking a variety of system and physiological metrics. In addition to survey questions, the system is capable of tracking the subject’s motions within the virtual environment, measuring variation in their grip strength (e.g., as an indicator of anxious arousal related to the crisis), and, via eyetracking, measuring shifts in visual attention and pupillometric indices of arousal. With respect to our research objectives related to robot-guided emergency evacuation, key metrics include whether the person follows the robot to (or away from) an exit, how long the person follows the robot, and how quickly the person evacuates. We also intend to assess information-foraging behavior before and during the crisis as indicated by head and eye movement, including assessments of whether and to what degree they noticed exit signs.  Although much of the work is exploratory, we hypothesize that the subjects will tend to follow the robot, even if the robot’s behavior is unreliable (i.e., increasing rather than decreasing risk of physical harm). We believe that evacuation time will be faster when a robot guides the evacuee to nearby exit, but slower when a robot erroneously directs evacuees away from nearby exits.  Assuming this prediction is borne out, we will analyze the aforementioned motor, eye-tracking and self-report data to identify reliable predictors of behavioral conformity with the robot.

\section{Upcoming Experiments}

Over the course of the next several months we intend to conduct a variety of experiments designed to test how people react to the emergency stimuli and the extent to which they are willing to follow a robot during an emergency. Assuming adequate human subject availability, we hope to investigate the impact that anthropomorphism has on the evacuee's decision to follow, how participants respond to different types of emergencies (active shooter, fire, ect), and validate the generalizability of the system to real-world dynamics by comparing experiments performed in VR to physical experiments performed in a real environment.     

\section{Conclusions}
This paper briefly describes our progress developing an experimental paradigm in virtual reality capable of evaluating robot guided emergency evacuation. Although we have only recently begun to conduct studies, we believe that the system will provide important insights towards understanding how and why people trust robots and choose to follow a robot during an emergency evacuation.     

\section{Acknowledgments}
This material is based upon work supported by the Air Force Office of Scientific Research under award number FA9550-20-1-0347.

\begin{small}
\bibliography{references}
\end{small}

\end{document}